\documentclass[11pt,a4paper]{article}
\usepackage[hyperref]{emnlp-ijcnlp-2019}
\usepackage{times}
\usepackage{latexsym}

\usepackage{url}
\usepackage{lipsum}

\usepackage{arydshln}
\usepackage{latexsym}
\usepackage{color}
\usepackage{array} 
\usepackage{booktabs}  
\usepackage{threeparttable}  
\usepackage{multicol}  
\usepackage{multirow}  
\usepackage{natbib}
\usepackage{epsfig}
\usepackage{algorithm}
\usepackage{algorithmic}
\usepackage{subcaption}
\aclfinalcopy 

\definecolor{white}{rgb}{1,1,1}

\definecolor{yellow}{rgb}{0.99,0.99,0.70}
\definecolor{black}{rgb}{0.00,0.00,0.00}

\newcommand*{\affaddr}[1]{#1} 
\newcommand*{\affmark}[1][*]{\textsuperscript{#1}}

\title{Induction Networks for Few-Shot Text Classification}

\author{
  Ruiying Geng\affmark[1,2], Binhua Li\affmark[2], Yongbin Li\affmark[2]$^*$ , Xiaodan Zhu\affmark[3], Ping Jian\affmark[1]\thanks{$^*$Corresponding authors: Y.Li and P.Jian.}\ , Jian Sun\affmark[2]\\
\affaddr{\affmark[1]School of Computer Science and Technology, Beijing Institute of Technology}\\
 \affaddr{\affmark[2] Alibaba Group, Beijing}\\
   \affaddr{\affmark[3] ECE, Queen's University}\\
{\tt \{ruiying.gry,binhua.lbh,shuide.lyb,jian.sun\}@alibaba-inc.com}\\
    {\tt zhu2048@gmail.com    }      {\tt pjian@bit.edu.cn} \\
  }

\date{}
\begin{document}
\maketitle

\begin{abstract}
Text classification tends to struggle when data is deficient or when it needs to adapt to unseen classes. In such challenging scenarios, recent studies have used meta-learning to simulate the few-shot task, in which new queries are compared to a small support set at the sample-wise level. However, this sample-wise comparison may be severely disturbed by the various expressions in the same class. Therefore, we should be able to learn a general representation of each class in the support set and then compare it to new queries. In this paper, we propose a novel Induction Network to learn such a generalized class-wise representation, by innovatively leveraging the dynamic routing algorithm in meta-learning. In this way, we find the model is able to induce and generalize better. We evaluate the proposed model on a well-studied sentiment classification dataset (English) and a real-world dialogue intent classification dataset (Chinese). Experiment results show that on both datasets, the proposed model significantly outperforms the existing state-of-the-art approaches, proving the effectiveness of class-wise generalization in few-shot text classification.
\end{abstract}

\section{Introduction}
Deep learning has achieved a great success in many fields such as computer vision, speech recognition and natural language processing \citep{kuang2018attention}. 
However, supervised deep learning is notoriously greedy for large labeled datasets, which limits the generalizability of deep models to new classes due to annotation cost. 
Humans on the other hand are readily capable of rapidly learning new classes of concepts with few examples or stimuli. This notable gap provides a fertile ground for further research.

Few-shot learning is devoted to resolving the data deficiency problem by recognizing novel classes from very few labeled examples. The limitation of only one or very few examples challenges the standard fine-tuning method in deep learning. Early studies \citep{salamon2017deep} applied data augmentation and regularization techniques to alleviate the overfitting problem caused by data sparseness, only to a limited extent. Instead, researchers have explored meta-learning \citep{finn2017model} to leverage the distribution over similar tasks, inspired by human learning. Contemporary approaches to few-shot learning often decompose the training procedure into an auxiliary meta-learning phase, which includes many meta-tasks, following the principle that the testing and training conditions must match. They extract some transferable knowledge by switching the meta-task from mini-batch to mini-batch. As such, few-shot models can classify new classes with just a small labeled support set.

However, existing approaches for few-shot learning still confront many important problems, including the imposed strong priors \citep{fei2006one}, complex gradient transfer between tasks \citep{munkhdalai2017meta}, and fine-tuning the target problem \citep{qi2018low}. The approaches proposed by \citet{snell2017prototypical} and \citet{sung2018learning}, which combine non-parametric methods and metric learning, provide potential solutions to some of those problems. The non-parametric methods allow novel examples to be rapidly assimilated, without suffering from catastrophic overfitting. Such non-parametric models only need to learn the representation of the samples and the metric measure. 
However, instances in the same class are interlinked and have their uniform fraction and their specific fractions. In previous studies, the class-level representations are calculated by simply summing or averaging representations of samples in the support set. In doing so, essential information may be lost in the noise brought by various forms of samples in the same class. Note that few-shot learning algorithms do not fine-tune on the support set. When increasing size of the support set, the improvement brought by a bigger data size will also be diminished by more sample level noises.

Instead, we explore a better approach by performing induction at the class-wise level: ignoring irrelevant details and encapsulating general semantic information from samples with various linguistic forms in the same class. 
As a result, there is a need for a perspective architecture that can reconstruct hierarchical representations of support sets and dynamically induce sample representations to class representations. 

Recently, capsule network \citep{sabour2017dynamic} has been proposed, which possesses the exciting potential to address the aforementioned issue. A capsule network uses ``capsules" that perform dynamic routing to encode the intrinsic spatial relationship between parts and whole that constitutes viewpoint invariant knowledge.
Following a similar spirit, we can regard samples as parts and class as a \textit{whole}.
We propose the Induction Networks, which aims to model the ability of learning generalized class-level representation from samples in a small support set, based on the dynamic routing process.
First, an Encoder Module generates representations for a query and support samples. 
Next, an Induction Module executes a dynamic routing procedure, in which the matrix transformation can be seen as a map from the sample space to the class space, and then the generation of the class representation is all depending on the routing-by-agreement procedure other than any parameters, which renders a robust induction ability to the proposed model to deal with unseen classes. 
By regarding the samples' representations as input capsules and the classes' as output capsules, we expect to recognize the semantics of classes that is invariant to sample-level noise. 
Finally, the interaction between queries and classes is modelled---their representations are compared by a Relation Module to determine if the query matches the class or not. Defining an episode-based meta-training strategy, the holistic model is meta-trained end-to-end with the generalizability and scalability to recognize unseen classes.

The specific contributions of our work are listed as follows:
\begin{itemize}
\item We propose the Induction Networks for few-shot text classification. To deal with sample-wise diversity in the few-shot learning task, our model is the first, to the best of our knowledge, that explicitly models the ability to induce class-level representations from small support sets.
\item The proposed Induction Module combines the dynamic routing algorithm with typical meta-learning frameworks. The matrix transformation and routing procedure enable our model to generalize well to recognize unseen classes. 
\item Our method outperforms the current state-of-the-art models on two few-shot text classification datasets, including a well-studied sentiment classification benchmark and a real-world dialogue intent classification dataset. 
\end{itemize}

\section{Related Work}
\subsection{Few-Shot Learning}
The seminal work on few-shot learning dates back to the early 2000s \citep{fe2003bayesian,fei2006one}. The authors combined generative models with complex iterative inference strategies. More recently, many approaches have used a meta-learning \citep{finn2017model,mishra2017simple} strategy in the sense that they extract some transferable knowledge from a set of auxiliary tasks, which then helps them to learn the target few-shot problem well without suffering from overfitting. In general, these approaches can be divided into two categories. 

\paragraph{Optimization-based Methods} This type of approach aims to learn to optimize the model parameters given the gradients computed from the few-shot examples. \citet{munkhdalai2017meta} proposed the Meta Network, which learnt the meta-level knowledge across tasks and shifted its inductive biases via fast parameterization for rapid generalization. \citet{mishra2017simple} introduced a generic meta-learning architecture called SNAIL which used a novel combination of temporal convolutions and soft attention.

\paragraph{Distance Metric Learning} These approaches are different from the above approaches that entail some complexity when learning the target few-shot problem. The core idea in metric-based few-shot learning is similar to nearest neighbours and kernel density estimation. The predicted probability over a set of known labels is a weighted sum of labels of support set samples. \citet{vinyals2016matching} produced a weighted K-nearest neighbour classifier measured by the cosine distance, which was called Matching Networks. \citet{snell2017prototypical} proposed the Prototypical Networks which learnt a metric space where classification could be performed by computing squared Euclidean distances to prototype representations of each class. Different from fixed metric measures, the Relation Network learnt a deep distance metric to compare the query with given examples \citep{sung2018learning}.

Recently, some studies have been presented focusing specifically on few-shot text classification problems. \citet{xu2018lifelong} studied lifelong domain word embeddings via meta-learning. \citet{yu2018diverse} argued that the optimal meta-model may vary across tasks, and they employed the multi-metric model by clustering the meta-tasks into several defined clusters. \citet{rios2018few} developed a few-shot text classification model for multi-label text classification where there was a known structure over the label space. \citet{xu2019open} proposed a open-world learning model to deal with the unseen classes in the product classification problem. We solve the few-shot learning problem from a different perspective and propose a dynamic routing induction method to encapsulate the abstract class representation from samples, achieving state-of-the-art performances on two datasets.

\subsection{Capsule Network}
The Capsule Network was first proposed by \citet{sabour2017dynamic}, which allowed the network to learn robustly the invariants in part-whole relationships. Lately, Capsule Network has been explored in the natural language processing field. \citet{zhao2018investigating} successfully applied Capsule Network to fully supervised text classification problem with large labeled datasets. Unlike their work, we study few-shot text classification. \citet{xia2018zero} reused the supervised model similar to that of \citet{zhao2018investigating} for intent classification, in which a capsule-based architecture is extended to compute similarity between the target intents and source intents. Unlike their work, we propose Induction Networks for few-shot learning, in which we propose to use capsules and dynamic routing to learn generalized class-level representation from samples based. The dynamic routing method makes our model generalize better in the few-shot text classification task. 





\section{Problem Definition}
\subsection{Few-Shot Classification}
Few-shot classification \citep{vinyals2016matching,snell2017prototypical} is a task in which a classifier must be adapted to accommodate new classes not seen in training, given only a few examples of each of these new classes. We have a large labeled training set with a set of classes $C_{train}$. However, after training, our ultimate goal is to produce classifiers on the testing set with a disjoint set of new classes $C_{test}$, for which only a small labeled support set will be available. If the support set contains $K$ labeled examples for each of the $C$ unique classes, the target few-shot problem is called a $C$-way $K$-shot problem. Usually, the $K$ is too small to train a supervised classification model. Therefore, we aim to perform meta-learning on the training set, and extract transferable knowledge that will allow us to deliver better few-shot learning on the support set and thus classify the test set more accurately.

\subsection{Training Procedure}
The training procedure has to be chosen carefully to match inference at test time. An effective way to exploit the training set is to decompose the training procedure into an auxiliary meta-learning phase and mimic the few-shot learning setting via episode-based training, as proposed in \citet{vinyals2016matching}. We construct an meta-episode to compute gradients and update our model in each training iteration. The meta-episode is formed by randomly selecting a subset of classes from the training set first, and then choosing a subset of examples within each selected class to act as the support set $S$ and a subset of the remaining examples to serve as the query set $Q$. 
The meta-training procedure explicitly learns to learn from the given support set $S$ to minimise a loss over the query set $Q$. We call this strategy as episode-based meta training, and the details are shown in Algorithm 1.
It is worth noting that there are exponentially many possible meta tasks to train the model on, making it hard to overfit. 
For example, if a dataset contains 159 training classes, this leads to $\left( \begin{array}{cc}
     159  \\
     5 
\end{array}\right)=794,747,031$ possible $5-way$ tasks.\\


\begin{algorithm}[t]
\caption{Episode-Based Meta Training}
\begin{algorithmic}[1]
\FOR{each $episode \; iteration$}
\STATE Randomly select $C$ classes from the class space of the training set;
\STATE Randomly select $K$ labeled samples from each of the $C$ classes as support set 
$S = \left\{ {\left( {{x_s},{y_s}} \right)} \right\}_{s = 1}^m\left( {m = K \times C} \right)$, and select a fraction of the reminder of those $C$ classes' samples as query set $Q = \left\{ {\left( {{x_q},{y_q}} \right)} \right\}_{q = 1}^n$;
\STATE Feed the support set $S$ to the model and update the parameters by minimizing the loss in the query set $Q$;
\ENDFOR
\label{episode based meta training_alg}
\end{algorithmic}
\end{algorithm}

\begin{figure*}[t]
\centering
\small
\includegraphics[width=13cm]{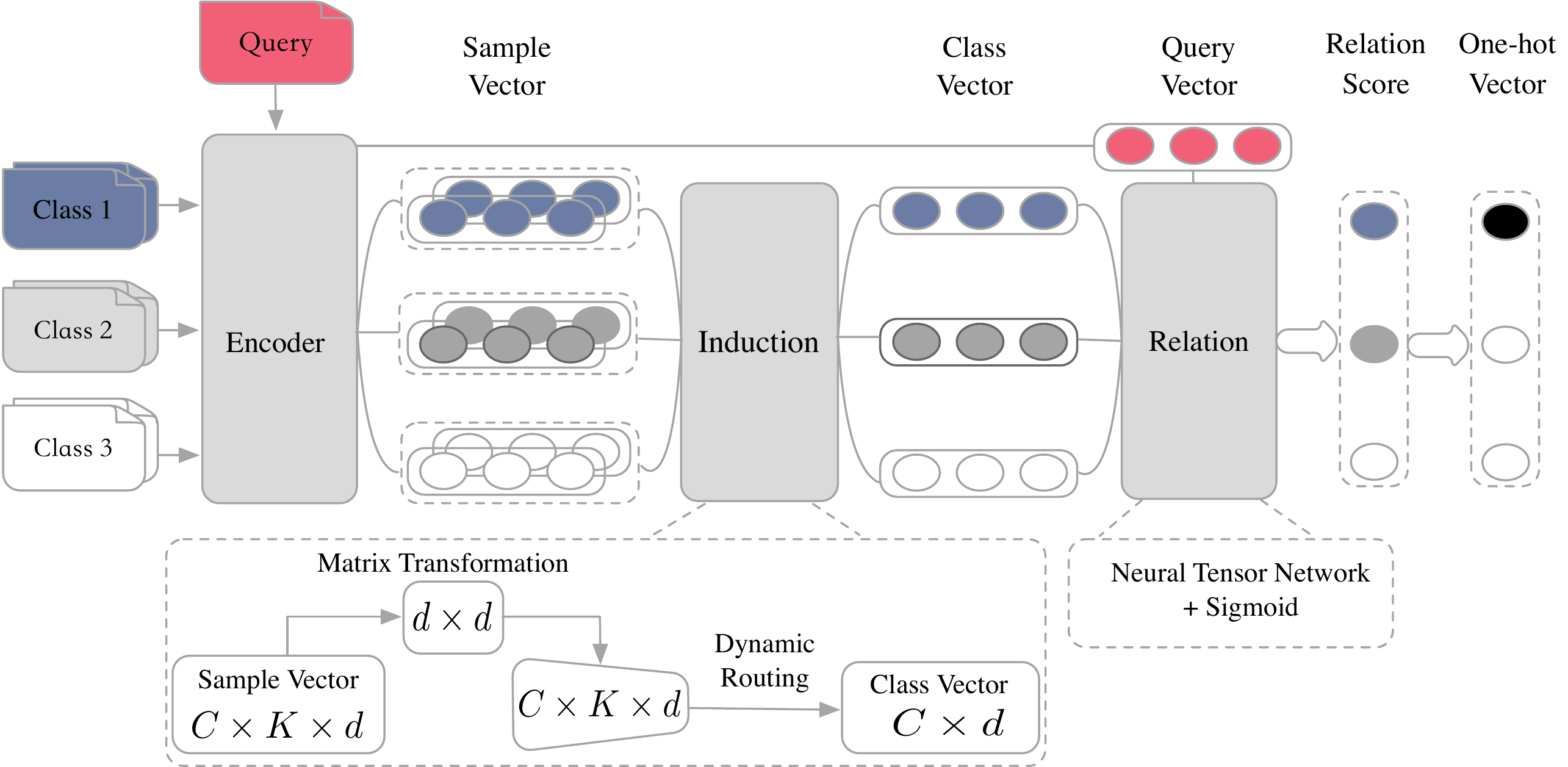}
\caption{Induction Networks architecture for a $C$-way $K$-shot ($C=3$, $K=2$) problem with one query example}
\centering
\label{model architechture}
\end{figure*}

\section{The Models}
Our Induction Networks, depicted in Figure \ref{model architechture} (the case of $3$-way $2$-shot model), consists of three modules: Encoder Module, Induction Module and Relation Module. In the rest of this section, we will show how these modules work in each meta-episode. 

\subsection{Encoder Module}
This module is a bi-direction recurrent neural network with self-attention as shown in \citet{lin2017structured}. Given an input text $x=(w_1,w_2,...,w_T)$, represented by a sequence of word embeddings. We use a bidirectional LSTM to process the text:
\begin{equation}
\quad\overrightarrow{h_t}=\quad\overrightarrow{LSTM}(w_t, h_{t-1})
\end{equation}
\begin{equation}
\quad\overleftarrow{h_t}=\quad\overleftarrow{LSTM}(w_t, h_{t+1})
\end{equation}
And we concatenate $\overrightarrow{h_t}$ with $\overleftarrow{h_t}$ to obtain a hidden state $h_t$. Let the hidden state size for each unidirectional LSTM be $u$.
For simplicity, we note all the $T\;h_ts$ as $H=(h_1, h_2,...,h_T)$. Our aim is to encode a variable length of text into a fixed size embedding. We achieve that by choosing a linear combination of the $T$ $LSTM$ hidden vectors in $H$. Computing the linear combination requires the self-attention mechanism, which takes the whole LSTM hidden states $H$ as input, and outputs a vector of weights $a$:
\begin{equation}
    a=softmax(W_{a2}tanh(W_{a1}H^T))
\end{equation}
here $W_{a1} \in R^{d_a \times 2u}$ and $W_{a2} \in R^{d_a}$ are weight matrices and $d_a$ is a hyperparameter. The final representation $e$ of the text is the weighted sum of $H$:
\begin{equation}
    e = \sum_{t=1}^T{a_t\cdot h_t}
    \label{encoder}
\end{equation}

\subsection{Induction Module}
This section introduces the proposed dynamic routing induction algorithm. We regard these vectors $e$ obtained from the support set $S$ by Eq~\ref{encoder} as sample vectors $e^s$, and the vectors $e$ from the query set $Q$ as query vectors $e^q$. The most important step is to extract the representation for each class in the support set. The main purpose of the induction module is to design a non-linear mapping from sample vector ${e}_{ij}^s$ to class vector $c_i$:\\
\[{\left\{ {{{e}_{ij}^s} \in {R^{2u}}} \right\}_{i = 1,...C,j = 1...K}} \mapsto \left\{ {{c_i} \in {R^{2u}}} \right\}_{i = 1}^C.\]

We apply the dynamic routing algorithm \citep{sabour2017dynamic} in this module, in the situation where the number of the output capsule is one. 
In order to accept $any$-way $any$-shot inputs in our model, a weight-sharable transformation across all sample vectors in the support set is employed. All of the sample vectors in the support set share the same
transformation weights ${W_s} \in {R^{2u \times 2u}}$ and bias $b_s$, so that the model is flexible enough to handle the support set at any scale. Each sample prediction vector $\hat{e}_{ij}^s$ is computed by:
\begin{equation}
{\hat{e}_{ij}^s} = squash({W_s}{{e}_{ij}^s} + b_s)
\label{squash_transformation}
\end{equation}
where $squash$ is a non-linear squashing function through the entire vector, which leaves the direction of the vector unchanged but decreases its magnitude. Given input vector $x$, $squash$ is defined as:
\begin{equation}
squash(x) = \frac{{{{\left\| {x} \right\|}^2}}}{{1 + {{\left\| {x} \right\|}^2}}}\frac{{x}}{{\left\| {x} \right\|}}
\end{equation}
Eq~\ref{squash_transformation} encodes important invariant semantic relationships between lower level sample features and higher level class features \citep{hinton2011transforming}.
To ensure the class vector encapsulates the sample feature vectors of this class automatically, dynamic routing is applied iteratively. In each iteration, the process dynamically amends the connection strength and makes sure that the coupling coefficients $d_i$ sum to $1$ between class $i$ and all support samples in this class by a ``routing softmax":
\begin{equation}
{d_{i}} = softmax \left( {b_{i}} \right)
\end{equation}
where $b_i$ is the logits of coupling coefficients, and initialized by $0$ in the first iteration. 
Given each sample prediction vector ${\hat{e}_{ij}^s}$, each class candidate vector ${\hat{c}_i}$ is a weighted sum of all sample prediction vectors ${\hat{e}_{ij}^s}$ in class $i$:
\begin{equation}
{\hat{c}_i} =  {\sum\limits_j {{d_{ij} \cdot {\hat{e}_{ij}^s}} }} 
\end{equation}
then a non-linear $``$squashing$"$ function is applied to ensure that the length of the vector output of the routing process will not exceed $1$:
\begin{equation}
{c_i} = squash({\hat{c}_i})
\label{squash}
\end{equation}

The last step in every iteration is to adjust the logits of coupling coefficients $b_{ij}$ by a ``routing by agreement" method. If the produced class candidate vector has a large scalar output with one sample prediction vector, there is a top-down feedback which increases the coupling coefficient for that sample and decreases it for other samples. This type of adjustment is very effective and robust for the few-shot learning scenario because it does not need to restore any parameters. Each $b_{ij}$ is updated by:
\begin{equation}
{b_{ij}} = {b_{ij}} + {\hat{e}_{ij}^s} \cdot {c_i}
\end{equation}

Formally, we call our induction method as dynamic routing induction and summarize it in Algorithm 2.
\begin{algorithm}[t]
\caption{Dynamic Routing Induction}
\begin{algorithmic}[1]
\REQUIRE sample vector ${e}_{ij}^s$ in support set $S$ and initialize the logits of coupling coefficients ${b_{ij}}=0$
\ENSURE class vector $c_i$
\STATE for all samples $j=1,...,K$ in class $i$:
\STATE $\;\;{\hat{e}_{ij}^s} = squash({W_s}{{e}_{ij}^s} + b_s)$
\FOR{ $iter$ iterations }
\STATE ${d_{i}} = softmax \left( {b_{i}} \right)$
\STATE ${\hat{c}_i} =  {\sum\limits_j {{d_{ij}} \cdot {\hat{e}_{ij}^s}} } $
\STATE ${c_i} = squash(\hat{c}_i)$
\STATE for all samples $j=1,...,K$ in class $i$:
\STATE $\quad{b_{ij}} = {b_{ij}} + {\hat{e}_{ij}^s} \cdot {c_i}$
\ENDFOR
\STATE \textbf{Return} $c_i$
\label{induction algorithm_alg}
\end{algorithmic}
\end{algorithm}

\subsection{Relation Module}
After the class vector $c_i$ is generated by the Induction Module and each query text in the query set is encoded to a query vector $e^q$ by the Encoder Module, the next essential procedure is to measure the correlation between each pair of query and class.
The output of the Relation Module is called the relation score, representing the correlation between $c_i$ and $e^q$, which is a scalar between $0$ and $1$. Specifically, we use the neural tensor layer \citep{socher2013reasoning} in this module, which has shown great advantages in modeling the relationship between two vectors \citep{wan2016deep,geng2017implicit}. We choose it as an interaction function in this paper. The tensor layer outputs a relation vector as follows:
\begin{equation}
v({c_i},{e^q}) = f\left( {c_i}^T{M^{[1:h]}}{e^q} \right)
\end{equation}
where ${M^k} \in {R^{2u \times 2u}},k \in \left[ {1,...,h} \right]$ is one slice of the tensor parameters and $f$ is a non-linear activation function called RELU \citep{glorot2011deep}. The final relation score $r_{iq}$ between the $i$-th class and the $q$-th query is calculated by a fully connected layer activated by a sigmoid function.
\begin{equation}
r_{iq}=sigmoid({W_r}v({c_i},{e^q})+b_r)
\end{equation}

\subsection{Objective Function}
We use the mean square error (MSE) loss to train our model, regressing the relation score $r_{iq}$ to the ground truth $y_q$: matched pairs have similarity $1$ and the mismatched pair have similarity $0$. Given the support set $S$ with $C$ classes and query set $Q=\left\{ {\left( {{x_q},{y_q}} \right)} \right\}_{q = 1}^n$ in an episode, the loss function is defined as:
\begin{equation}
L\left( {S, Q} \right) = \sum\limits_{i=1}^C {\sum\limits_{q=1}^n {{{({r_{iq}} - \textbf{1}({y_q}==i))}^2}} } \label{Loss}
\end{equation}
conceptually we are predicting relation scores, which can be considered as a regression problem and the ground truth is within the space $\left\{ {\left. {0,1} \right\}} \right.$.

All parameters of the three modules are trained jointly by backpropagation. The Adagrad \cite{duchi2011adaptive} is used on all parameters in each training episode. 
Our model does not need any fine-tuning on the classes it has never seen due to its generalization nature. The induction and comparison ability are accumulated in the model along with the training episodes.

\section{Experiments}
We evaluate our model by conducting experiments on two few-shot text classification datasets. All the experiments are implemented with Tensorflow.
\subsection{Datasets}
\paragraph{Amazon Review Sentiment Classification (ARSC)} Following \citet{yu2018diverse}, we use the multiple tasks with the multi-domain sentiment classification \citep{blitzer2007biographies} dataset. The dataset comprises English reviews for 23 types of products on Amazon. For each product domain, there are three different binary classification tasks. These buckets then form $23\times3=69$ tasks in total. Following \citet{yu2018diverse}, we select $12(4\times3)$ tasks from 4 domains (Books, DVD, Electronics and Kitchen) as the test set, and there are only five examples as support set for each label in the test set. We create $5$-shot learning models on this dataset. 
\begin{table} [t]
\centering
\small
\begin{tabular}{ccc}  
\toprule
 & \textbf{Training Set} & \textbf{Testing Set}\\ 
\midrule
Class Num& 159 & 57\\ 
Data Num & 195,775 & 2,279\\ 
Data Num/Class & $\ge 77$  & $20 \sim 77$\\ 
\bottomrule
\end{tabular}  
\caption{Details of ODIC}    
\label{table_odic}
\end{table}  

\paragraph{Open Domain Intent Classification for Dialog System (ODIC)} We create this dataset by fetching the log data on a real-world conversational platform. The enterprises submit various dialogue tasks with a great number of intents, but many intents have only a few labeled samples, which is a typical few-shot classification application. Following the definition of the few-shot learning task, we divide the ODIC into a training set and a testing set and ensure that the labels of the two sets have no intersection. The details of the set partition are shown in Table \ref{table_odic}.

\begin{table}[t]  
\centering
\small
\begin{tabular}{cc}  
\toprule
\textbf{Model}& \textbf{Mean Acc}\\ 
\midrule
Matching Networks \citep{vinyals2016matching}& 65.73  \\ 
Prototypical Networks \citep{snell2017prototypical} & 68.17\\ 
Graph Network \citep{Garcia2017FewShotLW} & 82.61\\ 
Relation Network \citep{sung2018learning} & 83.07\\  
SNAIL \citep{mishra2017simple} & 82.57\\  
ROBUSTTC-FSL \citep{yu2018diverse} & 83.12\\  
Induction Networks (ours) & \textbf{85.63} \\  
\bottomrule
\end{tabular}  
\caption{Comparison of mean accuracy (\%) on ARSC}
\label{ARSC results}
\end{table}  

\begin{table*}[t]  
\centering
\small
\begin{tabular}{cp{1.7cm}<{\centering}p{1.7cm}<{\centering}p{1.7cm}<{\centering}p{1.7cm}<{\centering}}  
\toprule
\multirow{2}*{\textbf{Model}} & \multicolumn{2}{c}{\textbf{5-way Acc.}} & \multicolumn{2}{c}{\textbf{10-way Acc.}}\\
~ & 5-shot & 10-shot & 5-shot & 10-shot \\
\midrule
Matching Networks \citep{vinyals2016matching} & 82.54$\pm$0.12 & 84.63$\pm$0.08 & 73.64$\pm$0.15 & 76.72$\pm$0.07 \\
Prototypical Networks \citep{snell2017prototypical} & 81.82$\pm$0.08& 85.83$\pm$0.06 & 73.31$\pm$0.14 & 75.97$\pm$0.11 \\
Graph Network \citep{Garcia2017FewShotLW} & 84.15$\pm$0.16& 87.24$\pm$0.09 & 75.58$\pm$0.12 & 78.27$\pm$0.10 \\
Relation Network \citep{sung2018learning} & 84.41$\pm$0.14 & 86.93$\pm$0.15 & 75.28$\pm$0.13 & 78.61$\pm$0.06 \\ 
SNAIL \citep{mishra2017simple} & 84.62$\pm$0.16& 87.31$\pm$0.11 & 75.74$\pm$0.07 & 79.26$\pm$0.09 \\ 
Induction Networks (ours) & \textbf{87.16}$\pm$0.09& \textbf{88.49}$\pm$0.17 & \textbf{78.27}$\pm$0.14 & \textbf{81.64}$\pm$0.08 \\  
\bottomrule
\end{tabular}  
\caption{Comparison of mean accuracy (\%) on ODIC}
\label{ODIC results}
\end{table*}

\subsection{Experiment Setup}
\paragraph{Baselines} In this section, the baseline models in our experiments are introduced as follows.
\begin{itemize}
\item Matching Networks: a few-shot learning model using a metric-based attention method \citep{vinyals2016matching}. 
\item Prototypical Networks: a deep metric-based method using sample average as class prototypes \citep{snell2017prototypical}.
\item Graph Network: a graph-based few-shot learning model that implements a task-driven message passing algorithm on the sample-wise level \citep{Garcia2017FewShotLW}.
\item Relation Network: a few-shot learning model which uses a neural network as the distance metric and sums up sample vectors in the support set as class vectors \citep{sung2018learning}.
\item SNAIL: a class of simple and generic meta-learner architectures that use a novel combination of temporal convolutions and soft attention \citep{mishra2017simple}.
\item ROBUSTTC-FSL: This approach combines several metric-based methods by clustering the tasks \citep{yu2018diverse}.
\end{itemize}

The baseline results on ARSC are reported in \citet{yu2018diverse} and we implemented the baseline models on ODIC with the same text encoder module.

\paragraph{Implementation Details} We use 300-dimension Glove embeddings \citep{pennington2014glove} for ARSC dataset and 300-dimension Chinese word embeddings trained by \citet{li2018analogical} for ODIC. We set the hidden state size of LSTM $u=128$ and the attention dimension $d_a=64$. The iteration number $iter$ used in dynamic routing algorithm is $3$. The relation module is a neural tensor layer with $ h=100 $ followed by a fully connected layer activated by sigmoid. We build $2$-way $5$-shot models on ARSC following \citet{yu2018diverse}, and build episode-based meta training with $ C=[5, 10]$ and $ K=[5, 10]$ for comparison on ODIC. In addition to $K$ sample texts as support set, the query set has 20 query texts for each of the $C$ sampled classes in every training episode. This means, for example, that there are $20\times5 + 5 \times 5 = 125$ texts in one training episode for the $5$-way $5$-shot experiments.

\paragraph{Evaluation Methods} We evaluate the performance by few-shot classification accuracy following previous studies in few-shot learning \citep{snell2017prototypical,sung2018learning}. To evaluate the proposed model with the baselines objectively, we compute mean few-shot classification accuracies on ODIC over 600 randomly selected episodes from the testing set. We sample 10 test texts per class in each episode for evaluation in both $5$-shot and $10$-shot scenarios. Note that for ARSC, the support set for testing is fixed by \citet{yu2018diverse}. Consequently, we just need to run the test episode once for each of the target tasks. The mean accuracy of the 12 target task is compared to the baseline models following \citet{yu2018diverse}.

\subsection{Experiment Results}

\paragraph{Overall Performance} Experiment results on ARSC are presented in Table \ref{ARSC results}. The proposed Induction Networks achieves a 85.63\% accuracy, outperforming the existing state-of-the-art model, ROBUSTTC-FSL, by a notable 3\% improvement. We due the improvement to the fact that ROBUSTTC-FSL builds a general metric method by integrating several metrics at the sample level, which faces the difficulty of getting rid of the noise among different expressions in the same class. In addition to that, the task-clustering-based method used by ROBUSTTC-FSL must be found on the relevance matrix, which is inefficient when applied to real-world scenarios where the tasks change rapidly. Our Induction Networks, however, is trained in the meta-learning framework with more flexible generalization and its induction ability can hence be accumulated through different tasks.

We also evaluate our method with a real-world intent classification dataset ODIC. The experiment results are listed in Table \ref{ODIC results}. We can see that our proposed Induction Networks achieves best classification performances on all of the four experiments. In the distance metric learning models (Matching Networks, Prototypical Networks, Graph Network and Relation Network), all the learning occurs in representing features and measuring distances at the sample-wise level. Our work builds an induction module focusing on the class-wise level of representation, which we claim to be more robust to variation of samples in the support set. Our model also outperforms the latest optimization-based method---SNAIL. The difference between Induction Networks and SNAIL shown in Table~\ref{ODIC results} is statistically significant under the paired at the 99\% significance level. 
In addition, the performance difference between our model and other baselines in the 10-shot scenario is more significant than in the 5-shot scenario. This is because in the 10-shot scenario, for the baseline models the improvement brought by a bigger data size is also diminished by more sample level noises.


\paragraph{Ablation Study} To analyze the effect of varying different components of the Induction Module and Relation Module, we further report the ablation experiments on the ARSC dataset as shown in Table~\ref{ablation}. We can see that the best performance is achieved when we used $3$ iterations, corresponding to the best result reported in Table~\ref{ARSC results} (more rounds of iterations did not further improve the performance), and the table shows the effectiveness of the routing component.
We also changed the Induction Module with sum and self-attention and changed Relation Module with cosine distance. Changes in the performances validate the benefit of both the Relation Module and Induction Module. The Attention+Relation models the induction ability by self-attention mechanism, but the ability is limited by the learnt attention parameters. Conversely, the proposed dynamic routing induction method captures class-level information by automatically adjusting the coupling coefficients according to inputted support sets, which is more suitable for the few-shot learning task.


\subsection{Further Analysis}
We further analyze the effect of transformation and visualize query text vectors to show the advantage of the Induction Networks.

\begin{table}[t]  
\centering
\small
\begin{tabular}{ccc}  
\toprule
& \textbf{Iteration}& \textbf{Accuracy}\\ 
\midrule
Routing+Relation& 3 & {\bf 85.63} \\  
Routing+Relation& 2 & 85.41\\  
Routing+Relation& 1 & 85.06 \\ 
\cdashline{1-3}[1pt/3pt]
Routing + Cosine & 3 & 84.67\\  
Sum +Relation& - & 83.07\\  
Attention + Relation & - & 85.15\\  
\bottomrule
\end{tabular}  
\caption{Ablation study of Induction Networks on ARSC dataset}
\label{ablation}
\end{table}  

\begin{figure}
    \centering
    \subcaptionbox{Before transformation
    \label{fig:subfig:a_trans}
    }{\includegraphics[width=3.5cm]{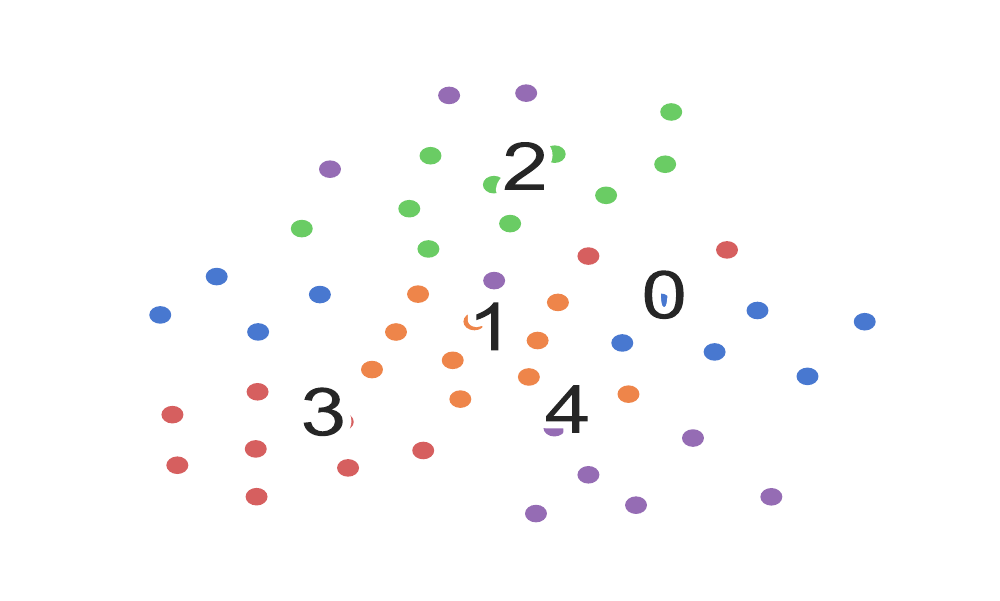}}
    \subcaptionbox{After transformation
    \label{fig:subfig:b_trans}
    }{\includegraphics[width=3.5cm]{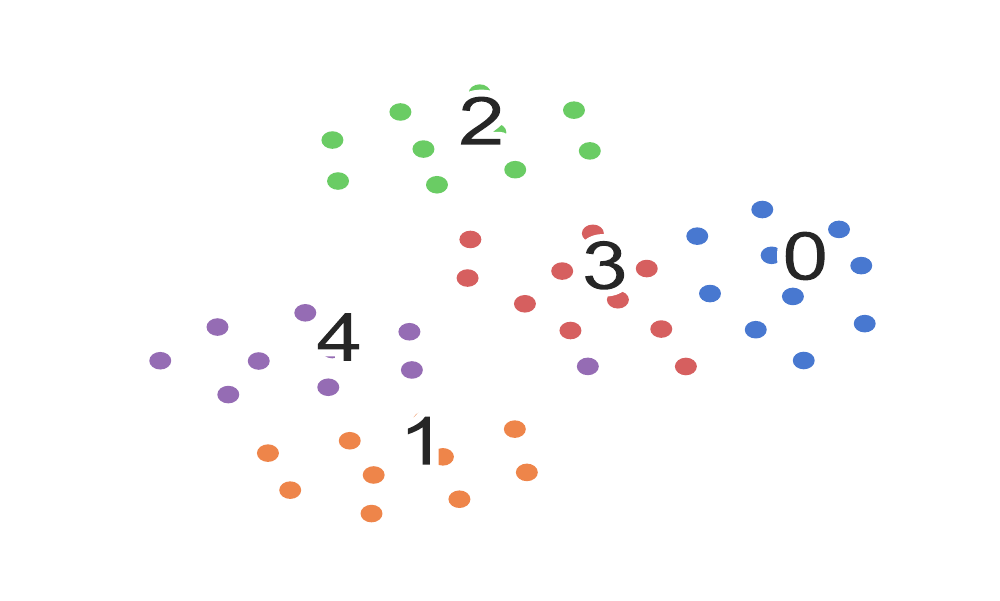}}
    \caption{Effect of Transformation under the $5$-way $10$-shot scenario. (a) The support sample vectors before matrix transformation. (b) The support sample vectors after matrix transformation.}
    \label{fig:transformation}
\end{figure}

\paragraph{Effect of Transformation}
Figure \ref{fig:transformation} shows the t-SNE \citep{maaten2008visualizing} visualization of support sample vectors before and after matrix transformation under the 5-way 10-shot scenario. We randomly select a support set with 50 texts (10 texts per class) from the ODIC testing set, and obtain the sample vectors ${\left\{ {{{e}_{ij}^s}} \right\}_{i = 1,...5,j = 1...10}}$ after the encoder module and the sample prediction vector ${\left\{ {{\hat{{e}}_{ij}^s}} \right\}_{i = 1,...5,j = 1...10}}$ after transformation. We can see that the vectors after matrix transformation are more separable, demonstrating the effectiveness of matrix transformation to encode semantic relationships between lower-level sample features and higher-level class features.

\begin{figure}
    \centering
    \subcaptionbox{Relation Network
    \label{fig:subfig:a}
    }{\includegraphics[width=3.5cm]{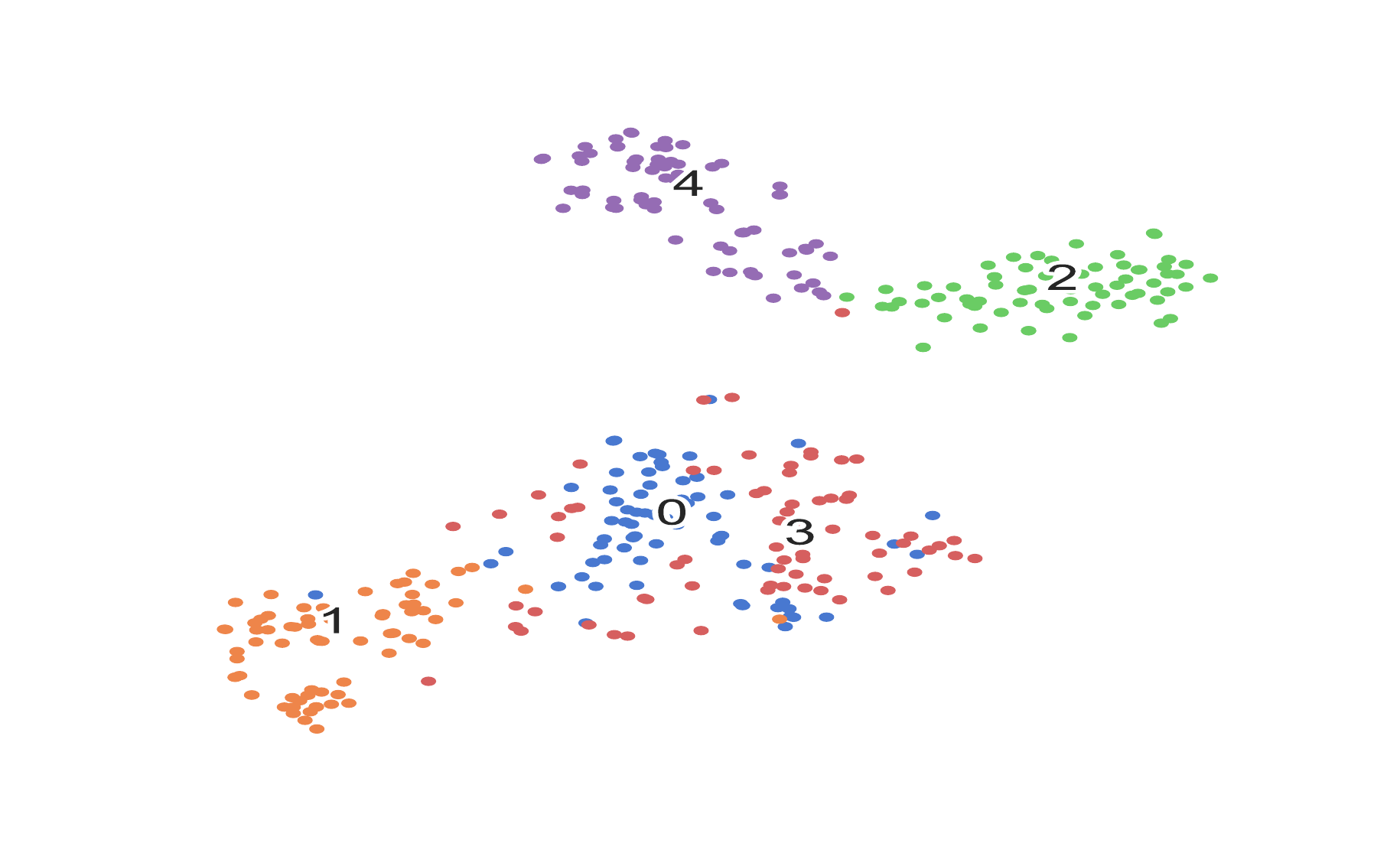}}
    \subcaptionbox{Induction Networks
    \label{fig:subfig:b}
    }{\includegraphics[width=3.5cm]{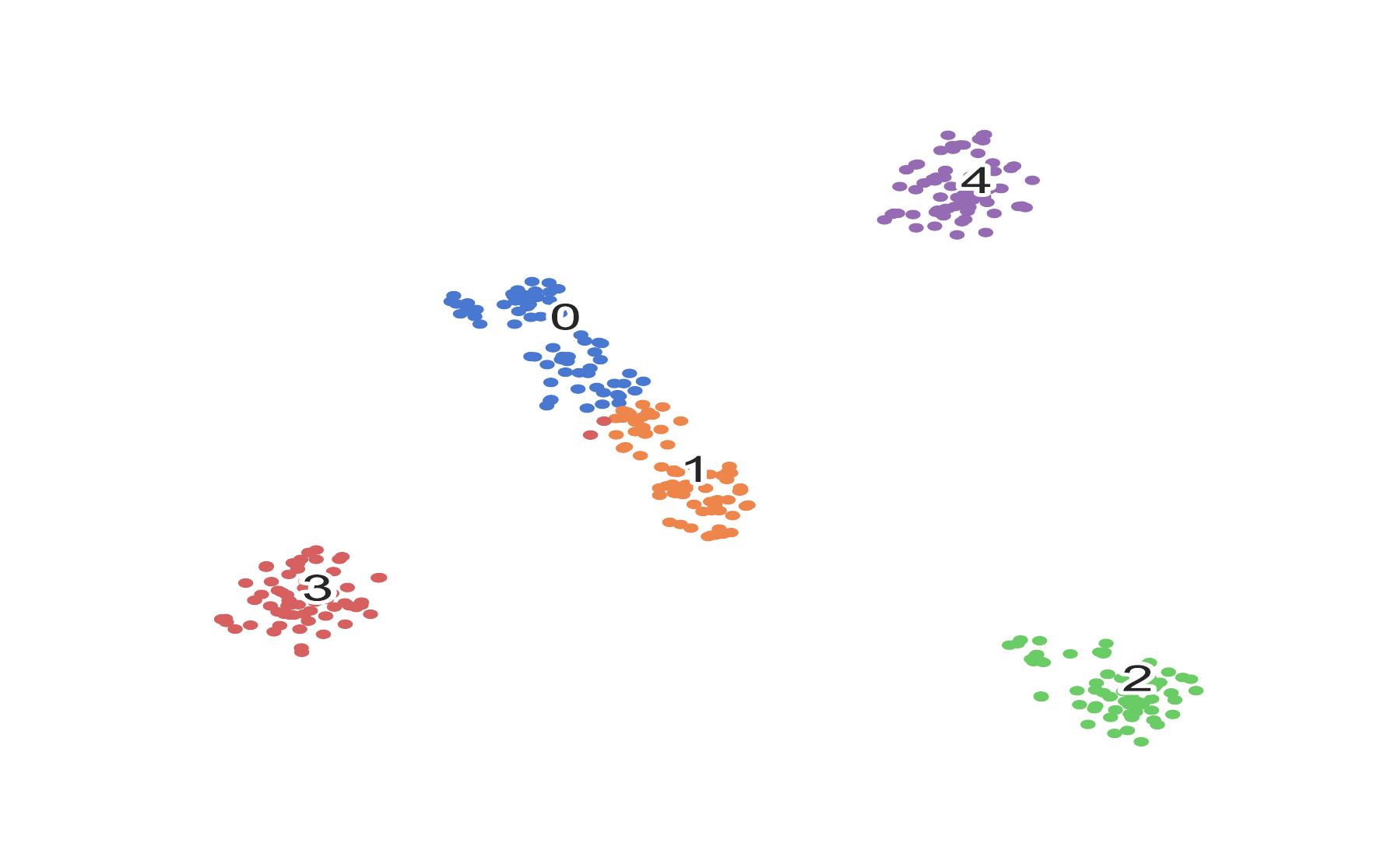}}
    \caption{Query text vector visualization learnt by (a) Relation Network and (b) Induction Networks.}
    \label{fig:sandian_cmp}
\end{figure}

\paragraph{Query Text Vector Visualization}
We also find out that our induction module does not only work well in generating effective class-level features, but also helps the encoders to learn better text vectors, as it can give different weights to instances and features during backpropagation. 
Figure \ref{fig:sandian_cmp} shows the t-SNE \citep{maaten2008visualizing} visualization of text vectors from the same randomly selected five classes, learnt by the Relation Network and our Induction Networks. 
It is clear that the text vectors learnt by Induction Networks are better separated semantically than those of Relation Network.

\section{Conclusion}
In this paper, we propose the Induction Networks, a novel neural model for few-shot text classification. We propose to induce the class-level representations from support sets to deal with sample-wise diversity in few-shot learning tasks. The Induction Module combines the dynamic routing algorithm with a meta-learning framework, and the routing mechanism makes our model more general to recognize unseen classes.
The experiment results show that the proposed model outperforms the existing state-of-the-art few-shot text classification models. We found that both the matrix transformation and routing procedure contribute consistently to the few-shot learning tasks.
\section*{Acknowledgments}
The authors would like to thank the organizers of EMNLP-IJCNLP2019 and the reviewers for their helpful suggestions. This research work is supported by the National Key Research and Development Program of China under Grant No. 2017YFB1002103, the National Natural Science Foundation of China under Grant No. 61751201, and the Research Foundation of Beijing Municipal Science \& Technology Commission under Grant No. Z181100008918002.

\bibliography{emnlp-ijcnlp-2019}
\bibliographystyle{acl_natbib}

\end{document}